\documentclass[pdflatex,sn-apa]{sn-jnl}% APA Reference Style
\usepackage{graphicx}%
\usepackage{multirow}%
\usepackage{amsmath,amssymb,amsfonts}%
\usepackage{amsthm}%
\usepackage{mathrsfs}%
\usepackage[title]{appendix}%
\usepackage{xcolor}%
\usepackage{textcomp}%
\usepackage{manyfoot}%
\usepackage{booktabs}%
\usepackage[linesnumbered,ruled,vlined]{algorithm2e}
\usepackage{algorithmicx}%
\usepackage{algpseudocode}%
\usepackage{listings}%
\usepackage{makecell}
\usepackage{tablefootnote}

\theoremstyle{thmstyleone}%
\theoremstyle{thmstyletwo}%

\theoremstyle{thmstylethree}%

\begin{document}

\title[Article Title]{Novel GPU Boruta algorithms for feature selection from high-dimensional data}


\author[1]{\fnm{Xurui} \sur{Li}}\email{lxr25@mails.tsinghua.edu.cn}

\author[1]{\fnm{Zhiguo} \sur{Gan}}\email{ganzhg22@163.com}

\author[1]{\fnm{Jiaming} \sur{Zhang}}\email{18801027274@163.com}
\author[1]{\fnm{Zheng} \sur{Liu}}\email{liuzheng@tsinghua.edu.cn}
\author*[1]{\fnm{Diannan} \sur{Lu}}\email{ludiannan@tsinghua.edu.cn}
% Department of Chemical Engineering, Tsinghua University, Beijing 100084, People's Republic of China
\affil[1]{\orgdiv{Department of Chemical Engineering}, \orgname{Tsinghua University}, \orgaddress{\street{Shuangqing Road}, \city{Beijing}, \postcode{100084}, \country{China}}}

% \affil[2]{\orgdiv{Department}, \orgname{Organization}, \orgaddress{\street{Street}, \city{City}, \postcode{10587}, \state{State}, \country{Country}}}

% \affil[3]{\orgdiv{Department}, \orgname{Organization}, \orgaddress{\street{Street}, \city{City}, \postcode{610101}, \state{State}, \country{Country}}}


\abstract{
    Most feature selection algorithms, especially wrapper methods, run inefficiently on CPU based platforms because of their high computational complexity. This inefficiency makes them unsuitable for processing large scale datasets. To address this challenge, the present study proposed two GPU accelerated versions of the Boruta feature selection procedure, in which Boruta-Permut relies on permutation based feature importance and Boruta-TreeImp employs importance based on impurity reduction.  To evaluate these methods we conducted experiments on both a self constructed dataset and several publicly available datasets. The experimental results show that the proposed GPU accelerated algorithms greatly improve computational efficiency while preserving feature selection accuracy comparable to the original Boruta algorithm. In our analysis we also observe that the impurity reduction based version can overestimate the importance of some features. Overall these findings suggest that performing Boruta feature selection on GPUs offers an effective and cost efficient solution for large scale data analysis, which is a good deal.
    }

\keywords{Machine Learning; GPU acceleration; Feature Selection; Feature Ranking; Random Forest}



\maketitle

\section{Introduction}
Recent years have witnessed growing efforts on applying artificial intelligence (AI) and machine learning (ML) on bioinformatics, pharmaceutical sciences, computational chemistry, and materials discovery  (\cite{Janet2020,FU2025,Zhang2025}).
However, scientific data in these fields often exhibit high dimensionality, strong nonlinearity, and complex inter-feature correlations.
While missing features leads to information loss and thus degrades model performance, redundant features may introduce noise, cause overfitting, and reduce generalization ability (\cite{Barraza2021}).
Feature engineering presents a grand challenge for developing efficient and robust artificial intelligence for science (AI4S) models.

Currently, two primary approaches are used in AI4S feature engineering to reduce the number of features: feature extraction and feature selection (\cite{Venkatesh2019}).
Feature extraction reduces dimensionality by transforming the original features into a new, lower-dimensional feature space. Common methods include Principal Component Analysis (PCA) (\cite{Gewers2021}), and otherwise such as Linear Discriminant Analysis (LDA), Normal Discriminant Analysis (NDA), and Canonical Variates Analysis (CVA) based on Fisher's linear discriminant (\cite{Fisher1936}).
However, at the fields of AI4S, the transformed features often lack physical meaning, which lowers the interpretability of the ML models. In contrast, feature selection reduces the feature set by identifying and retaining the most relevant original features, preserving their inherent meaning. 
% 把such cases补充一下
Hence, feature selection is the preferred approach (\cite{Li2017,Sebasti_n_2024}).

% Feature selection methods, however, must balance computational complexity and selection effectiveness when handling high-dimensional data.
The feature selection methods can be grouped into three categories: Filter method, Wrapper method, and Embedded method (\cite{Guyon2003,Guyon2006,Pudjihartono2022}).
\begin{enumerate}
    \item Filter method ranks or filters features by evaluating their individual statistical properties, such as correlation, mutual information, or variance. This method is computationally efficient, processes large datasets quickly, and helps reduce the risk of overfitting (\cite{Pudjihartono2022}). However, it treats each variable independently, ignores dependencies between features, and cannot resolve multicollinearity (\cite{Krakovska2019}), making it unsuitable for AI4S scenarios with many complex features.
    \item Embedded method integrates feature selection into the model training process, seeking features that are beneficial for specific steps (\cite{ElDahshan2023}). For example, Least Absolute Shrinkage and Selection Operator (LASSO) regression uses L1 regularization to shrink feature coefficients, thereby performing feature selection. % Although Embedded method is more efficient than Wrapper method, it still faces challenges in terms of generalizability due to its tight coupling with the model training process. However, only a few algorithms support the efficient implementation of the Embedded method, which limits its use.
    Although more efficient than the Wrapper method, the Embedded method's outputs are highly model-specific and lack generalizability across different models (\cite{Patel2025}).
    %it still faces scalability challenges with ultra-high-dimensional or massive-sample datasets.写明Embedded method的问题（例如……）
    \item Wrapper method transforms feature selection into a search problem. It uses a predictive model to evaluate the performance of feature subsets, iteratively adding or removing features to find the optimal subset. This approach considers feature interactions and, through its reliance on the ML model, ensures the selected subset benefits the target algorithm (\cite{Njoku2023}). However, its search space grows exponentially, and the iterative computations are costly (\cite{GONZALEZ2019}), limiting its use on large-scale datasets.
\end{enumerate}

The Boruta algorithm is a CPU-based, random forest-based Wrapper method for feature selection, which is developed by \cite{Kursa2010}.
Over the years, various studies have demonstrated its effectiveness across different feature sets (\cite{Degenhardt2019,Handhika2023,Gasmi2023,Yutika2023}).
However, the original paper also pointed out that the computation time of Boruta is proportional to $\text{attributes}\times\text{objects}$ (\cite{Kursa2010}), leading to inefficiency when processing high-dimensional datasets.

%% However, the original paper also pointed out that the computation time of Boruta is proportional to $n\times m$ (samples × features) or ($\text{attributes}\times\text{objects}$)  (\cite{Kursa2010}), leading to inefficiency when processing high-dimensional datasets.

To address this, we propose two GPU-accelerated variants of the Boruta algorithm, namely Boruta-Permut and Boruta-TreeImp. Both algorithms leverage the high parallelism of GPUs to accelerate the computation of feature importance, thereby speeding up the overall execution of Boruta.
We evaluated these two methods on both a self-constructed dataset and public datasets, using the Boruta implementation in scikit-learn and the BorutaShap algorithm (\cite{BorutaShap}) as references. The results show that these two algorithms maintain selection effectiveness while significantly improving computational efficiency.

\section{Methodology}
Boruta (\cite{Kursa2010}) is a feature selection algorithm based on random forests. It identifies which features are significantly important by comparing the importance of the original features against that of their randomized versions, known as ``shadow features''.
Specifically, the main steps of this algorithm are shown in Algorithm~\ref{alg:Boruta}.

\begin{algorithm}[htbp]
\caption{Pseudocode of the Boruta Algorithm}
\label{alg:Boruta}
\KwIn{Dataset $\mathcal{D} = \{(X,y)\mid X\in \mathbb{R}^{n\times p},\,y\in \mathbb{R}^n\}$, maximum iterations $t_{\max}$}
\KwOut{Feature importance ranking $\textit{rank}$, feature accept/reject state $\textit{state}$}
% initialization
$t \leftarrow 0$\;
$\textit{state} \leftarrow \mathbf 0_p$ \tcp*{feature state: 0 undecided, 1 accepted, -1 rejected}
$hit \leftarrow \mathbf 0_p$   \tcp*{count of times original feature beats shadow features}
\While{$t < t_{\max}$ \textbf{and} $\#\{i\mid \textit{state}_i=0\} > 0$}{
    $t \leftarrow t + 1$\;
    $X_{\text{cur}} \leftarrow X[:,\,\textit{state}\ge0]$\tcp*{columns of undecided features}
    $X_{\text{sha}} \leftarrow \text{copy}(X_{\text{cur}})$\tcp*{duplicate as shadow features}
  
    % ensure at least 5 shadow features
    \While{$\text{ncol}(X_{\text{sha}}) < 5$}{
        $X_{\text{sha}} \leftarrow X_{\text{sha}} \,\Vert\, X_{\text{sha}}$
        \tcp*{"$\Vert$" denotes column-wise concatenation}
    }
    $X_{\text{sha}} \leftarrow \text{shuffle}(X_{\text{sha}},\,\text{axis}=0)$
    \tcp*{shuffle shadow features by rows}
    % train model and compute importance
    $\mathcal{M} \leftarrow \text{fit}\bigl([X_{\text{cur}}\;\Vert\;X_{\text{sha}}],\,y\bigr)$\\
    $(\mathcal{I}_{\text{cur}},\,\mathcal{I}_{\text{sha}}) \leftarrow \text{Imp}(\mathcal{M})$
    \tcp*{extract importance of original and shadow features}
    % accumulate historical importance and hits
    $\mathcal{I}_{\text{his}} \leftarrow \mathcal{I}_{\text{his}} \,\Vert\, \mathcal{I}_{\text{cur}}$\;
    $hit \leftarrow hit + \mathbf{1}\bigl[\mathcal{I}_{\text{cur}} > \max(\mathcal{I}_{\text{sha}})\bigr]$\;
    % update state by significance test
    $\textit{state} \leftarrow \text{test}(\textit{state},\,hit,\,t)$
    \tcp*{state=1 if hit count is significant, state=-1 otherwise; others stay 0}
}
% final ranking
$rank \leftarrow \mathbf{1}_p$\;
$rank[\textit{state}=0] \leftarrow 2$\tcp*{assign medium importance to undecided features}
$rank[\textit{state}=-1] \leftarrow \text{rankdata}(\mathcal{I}_{\text{his}},\,\textit{state})$
\tcp*{sort rejected features by descending historical importance}
\Return{$\textit{rank},\;\textit{state}$}\;
\end{algorithm}

In the original paper, Kursa et al. measured feature importance by computing the decrease in classification accuracy resulting from randomly permuting feature values among samples. In scikit-learn  (\cite{sklearn_fi_demo}), feature importance is typically assessed by accumulating the impurity decrease within each tree and then computing the mean and standard deviation of these accumulated values.

Based on this, we use the Random Forest in cuML (\cite{raschka2020machine}) and implement the two Boruta algorithms. 
% 说明要清楚一点
Both algorithms follow the same idea of the original Boruta algorithm; however, by executing on the GPU, they achieve higher computational efficiency compared to the original implementations.

\subsection{GPU-Accelerated Boruta-TreeImp Algorithm}

In the Boruta-TreeImp algorithm, feature importance is quantified by the total impurity reduction of each feature across all trees. Specifically, we accelerate Random Forest construction with cuML and retrieve the structure of each tree; the feature importance calculation for each tree is shown in Algorithm~\ref{alg:nodeimportance} which run on CPU.
In essence, the Boruta-TreeImp algorithm implemented in this study can be viewed as a GPU implementation of the original impurity-based feature importance in scikit-learn.
% 因此本质上说而言，本研究实现的Boruta-TreeImp算法可视为对scikit-learn中原始impurity-based feature importance的GPU实现。

\begin{algorithm}[htbp]
\caption{Calculate Feature Importances via Impurity Decrease}
\label{alg:nodeimportance}
\KwIn{List of decision trees $nodes$, number of features $n$}
\KwOut{Vector of normalized average feature importance $I \in \mathbb{R}^{n}$}
\BlankLine
$imp \leftarrow \mathbf{0}_{|nodes|\times n}$\;
$feature\_gains \leftarrow \mathbf{0}_{n}$\;
\BlankLine
\SetKwFunction{CalcNodeImp}{CalcNodeImp}   % 注册函数名
\SetKwProg{Fn}{Function}{:}{end}
\Fn{\CalcNodeImp{$node$}}{
    $samples \leftarrow node.instance\_count$ (default $0$)\;
    $gain \leftarrow node.gain$\;
    $feature \leftarrow node.split\_feature$\;
    $feature\_gains[feature] \leftarrow feature\_gains[feature] + gain \times samples$\;
    $childs \leftarrow node.children$ (if exists)\;
    \ForEach{$child$ in $childs$}{
        \If{$child$ has attribute {\tt gain}}{
            \CalcNodeImp{$child$}\;
        }
    }
}
\BlankLine
\For{$i \leftarrow 0$ \KwTo $|nodes|-1$}{
    \CalcNodeImp{$nodes[i]$}\;
    $imp[i,\,\cdot] \leftarrow feature\_gains / \bigl(\sum_{j=0}^{n-1} feature\_gains[j]\bigr)$\;
    $feature\_gains \leftarrow \mathbf{0}_{n}$\;
}
$imp \leftarrow \frac{1}{|nodes|}\sum_{i=0}^{|nodes|-1} imp[i,\,\cdot]$\;
% $I \leftarrow imp / \bigl(\sum_{j=0}^{n-1} imp[j]\bigr)$\;
\Return{$imp / \bigl(\sum_{j=0}^{n-1} imp[j]\bigr)$}\;
\end{algorithm}

\subsection{GPU-Accelerated Boruta-Permut Algorithm}
Based on the original Boruta paper, we implement a GPU-based, permutation-invariance feature importance computation method.
This method measures the importance of each feature by randomly shuffling its values and computing the resulting change in the model's predictions.
For regression tasks, we use mean squared error (MSE) as the loss function; for classification tasks, we use the predicted class probabilities and apply the KL divergence (\cite{Kullback1951}) as the loss function. The detailed procedure is shown in Algorithm~\ref{alg:Permutationimportance}.

\begin{algorithm}[htbp]
\caption{Feature Importance via Permutation}
\label{alg:Permutationimportance}
\KwIn{Data matrix $X\in\mathbb{R}^{n\times p}$, target vector $y\in\mathbb{R}^n$, fitted estimator object}
\KwOut{Importance scores $\textit{imp}\in\mathbb{R}^p$}
% train and prepare model
% estimator.fit(X,y)\;
$\mathcal{M} \leftarrow$ fit(X,y)\;
% move data to GPU and get baseline predictions
$X_{\text{gpu}}\leftarrow\text{cp.asarray}(X)$\;
$y_{\text{gpu}}\leftarrow\mathcal{M}\text{.predict}(X_{\text{gpu}})$\;
% initialize structures
$p\leftarrow\text{ncol}(X_{\text{gpu}})$\;
$\text{imp}\leftarrow\mathbf{0}_{p}$\;
$\text{flush\_index}\leftarrow\text{random.permutation}(n)$\;
$X_{\text{work}}\leftarrow X_{\text{gpu}}.\text{copy}()$\;
$\text{original\_col}\leftarrow\mathbf{0}_{n}$\;
% permutation loop over features
\For{$i=0$ \KwTo $p-1$}{
    % save and replace column
    $\text{original\_col}[:] \leftarrow X_{\text{work}}[:,i]$\;
    $X_{\text{work}}[:,i] \leftarrow X_{\text{gpu}}[\text{flush\_index},i]$\;
    % predict and compute MSE
    $y_{\text{pred}}\leftarrow \mathcal{M}\text{.predict}(X_{\text{work}})$\;
    $\text{imp}[i]\leftarrow \text{LossFunction}(y_{\text{gpu}},\,y_{\text{pred}})$\;
    % restore column
    $X_{\text{work}}[:,i]\leftarrow \text{original\_col}$\;
}
\Return{imp}\;
\end{algorithm}

\subsection{Computational Complexity Analysis}
Assuming the dataset has $n$ samples and $m$ features, and the random forest model consists of $M$ trees, each with a average depth of $d$, each tree is approximately a full binary tree, so the number of leaf nodes is approximately $2^d$.

For Random Forest training, the time complexity could be approximated as (\cite{Hassine2019}):
\begin{equation}
    RF_{\textit{Tcomplexity}} = \mathcal{O}(M \cdot m \cdot n \cdot \log n)
\end{equation}

For origin Boruta algorithm which is in scikit-learn, the time complexity is mainly due to training the random forest and calculating feature importance for each iteration.
Calculating feature importance involves traversing each tree, which has a time complexity of $\mathcal{O}(M \cdot 2^d)$.
Therefore, the overall time complexity of the original Boruta algorithm is:
\begin{equation}
    \mathcal{O}(M \cdot m \cdot n \cdot \log n + M \cdot 2^d)
\end{equation}
Although it can be run on multiple CPU cores, it is still inefficient for high-dimensional datasets.

For the Boruta-TreeImp algorithm, the time complexity also includes training the random forest and calculating feature importance by traversing each tree.
Thus, the overall time complexity of the Boruta-TreeImp algorithm is:
\begin{equation}
    \mathcal{O}(M \cdot m \cdot n \cdot \log n + M \cdot 2^d)
\end{equation}
However, since its tree-building process is parallelized on the GPU, it is more efficient when handling high-dimensional datasets.
The main bottleneck lies in the traversal of decision trees, especially when the depth of the trees is large.

For the Boruta-Permut algorithm, the time complexity also includes training the random forest and computing feature importance through permutation.
For predicting a single sample, the time complexity is $\mathcal{O}(M \cdot m)$ (\cite{Hassine2019}), so predicting all samples has a time complexity of $\mathcal{O}(M \cdot n \cdot m)$. Considering there are $m$ features, requiring $m$ predictions, the total time complexity is:
\begin{equation}
    \mathcal{O}(M \cdot m \cdot n \cdot \log n + M \cdot n \cdot m^2)
\end{equation}

In summary, both algorithms have a time complexity dominated by the random forest training process, with additional costs for feature importance calculation. The Boruta-TreeImp algorithm is generally more efficient when the number of features $m$ is large, while the Boruta-Permut algorithm may be more suitable when $m$ is relatively small.

\section{Results and Discussion}
To evaluate the effectiveness of two new selection algorithms, we conducted experiments on a self-built dataset and publicly available datasets.
The experiments were run on a computer equipped with an AMD EPYC 7H12 64-Core Processor and an NVIDIA RTX 3090. In CPU-based runs, each algorithm used 128 cores; in GPU-based runs, each algorithm used one CPU core and a single GPU.
The reference prices for the CPU and GPU servers are listed in Table \ref{tab:server_price}.
\begin{table}[!htb]
    \centering
    \caption{Server Rental Price}
    \label{tab:server_price}
    \begin{tabular}{llll}
        \toprule
        Server Type & Hardware & Price$^*$ & Data Source \\
        \midrule
        CPU & \makecell[l]{AMD EPYC 7H12(128cores)\\ecs.g6a.32xlarge} & 4.881 USD/hr & \cite{AlibabaCloud} \\
        GPU & NVIDIA RTX 3090 & 0.118 USD/hr & \cite{VastAI} \\
        \bottomrule
    \end{tabular}
    $^*$All prices are the lowest available as of September 21, 2025.
\end{table}

\subsection{Synthetic experiments}
We selected common elementary functions to construct the following self-built dataset:
\begin{equation}
    \begin{aligned}
        f\left(\boldsymbol{X}\right) = &5X_{27}^{3} + 4X_{49}^{2} + 5X_{31}X_{46}  + 2X_{14} - 2.5X_{40} + 3.5\sqrt[3]{X_{18}} \\
        &+ e^{X_{20}} + 2\sin(3.14X_{26}) + 5\cos(3.14X_{33}) + \epsilon_1
    \end{aligned}
\end{equation}

Here, $\boldsymbol{X} = \left(X_1, X_2, \ldots, X_{50}\right)$; only 10 variables were used to generate the target value, and the other 40 variables were pure noise. We sampled $X_i \sim \mathcal{U}(0, 1)$ and $\epsilon_1 \sim \mathcal{N}(0, 0.00001)$.
We randomly generated 10,000 samples and used \verb|RandomForestRegressor| as the base learner. Feature selection was performed using the original Boruta algorithm, the Boruta-Permut, and the Boruta-TreeImp algorithms. For each method, we ran 20 experiments with different random seeds, recorded the feature importance from each run, and finally used the median importance as the basis for the final selection.
% \subsubsection{Direct Selection Results}
% We directly applied the three algorithms to the dataset; the results are shown in Table \ref{tab:simu_direct1}. Both Boruta and Boruta-Permut correctly selected all 10 important features. Boruta-TreeImp, due to its tree-structure approximation for importance calculation, failed to select feature number 18 but did not include any additional noise features.

\subsubsection{Direct Screening Results}
Three algorithms were directly applied to the dataset, and the results are shown in Table~\ref{tab:simu_direct1}.
It can be seen that both Boruta and Boruta-Permut correctly selected all 10 important features, whereas Boruta-TreeImp, due to its tree-based approximation of feature importance, failed to select feature-18 but did not include any additional noise features.

\begin{figure}[!htb]
    \centering
    \includegraphics[width=\textwidth]{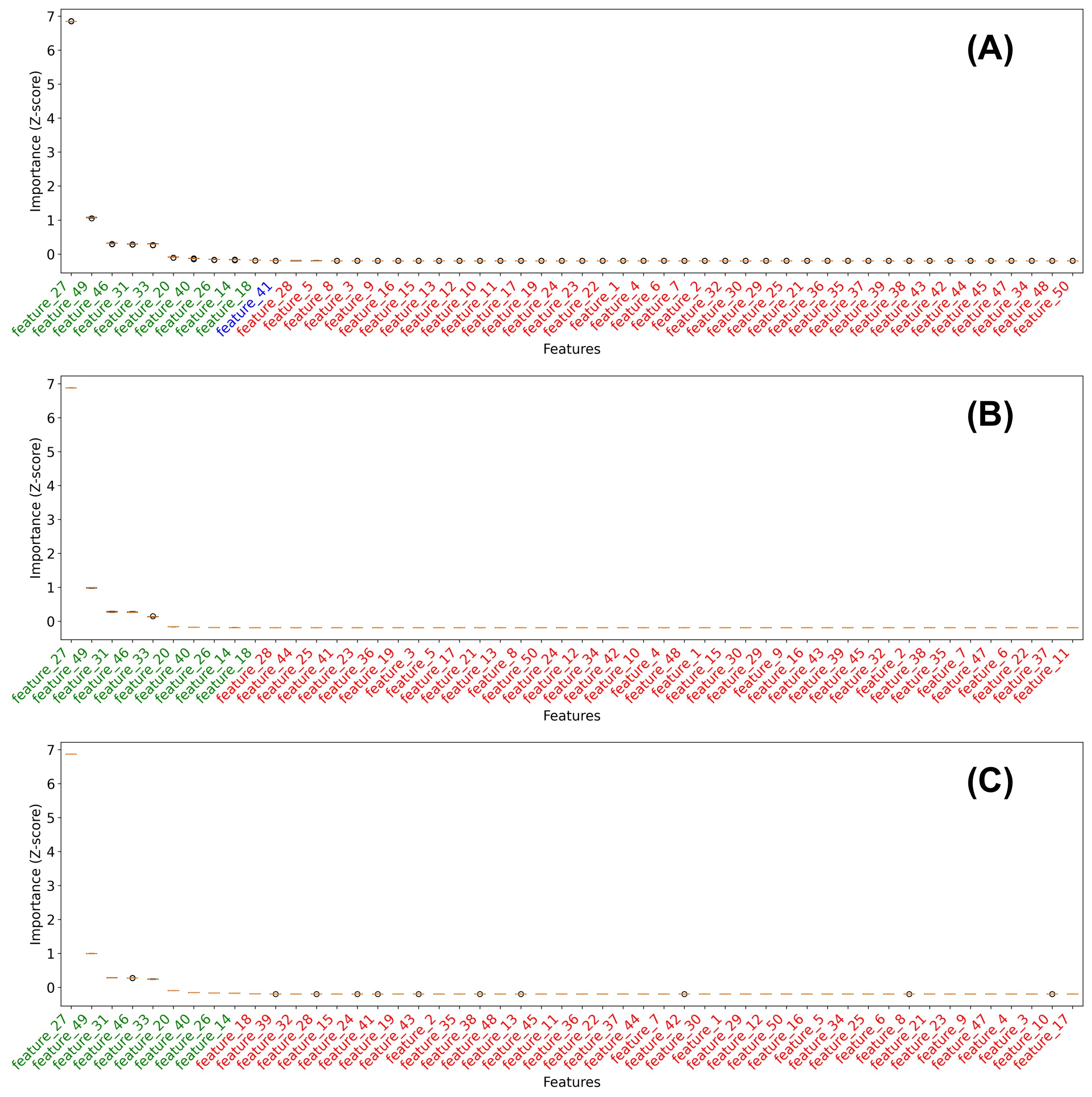}
    \caption{
        \textbf{Comparison of Direct Selection Results on the Self-Constructed Dataset}
        \\
        (A) Boruta; (B) Boruta-Permut; (C) Boruta-TreeImp.
        \\
        Green shows accepted features, blue shows tentative features, red shows rejected features
    }
    \label{fig:simu_direct1}
\end{figure}

As shown in Figure~\ref{fig:simu_direct1}, all three algorithms effectively distinguish important features from noise features, and the ranking of the important features is largely consistent.
It can be seen that, although the Boruta-TreeImp placed the feature-18 in a relatively high position, it did not select it. This finding is consistent with the official scikit-learn conclusion that impurity-based feature importance may overestimate the importance of some random features  (\cite{sklearnPermImp2025}). Considering that the feature-18 enters the objective function as a cube root and ranges from 0 to 2, its actual contribution is small, while some shadow features have inflated importance. Therefore, the Boruta-TreeImp did not select this feature.

% 可以看到Boruta-TreeImp虽然将第18个特征排在了较前的位置，但未能成功选出。
% 这符合scikit-learn官方给出的impurity-based feature importance可能会夸大某些随机特征重要性的结论 (\cite{sklearnPermImp2025})。
% 考虑到第18个特征在目标函数中以三次方根的形式出现，且取值范围为0到2，导致其实际贡献较小，同时部分Shadow特征重要性被高估，因此未能被Boruta-TreeImp选出。
\begin{table}[!htb]
    \centering
    \caption{Comparison of Direct Selection Results on the Self-Constructed Dataset}
    \label{tab:simu_direct1}
    \begin{tabular}{llllll}
        \toprule
        Algorithm & Time & Average Cost & № Feats & Unselected & Additional \\
        \midrule
        Boruta & 19m33s & 1.5904 USD & 10 & / & / \\
        Boruta-Permut & \textbf{03m28s} & \textbf{0.0068 USD} & 10 & / & / \\
        Boruta-TreeImp & 04m41s & 0.0092 USD & 9 & feature-18 & / \\
        \bottomrule
    \end{tabular}
    \begin{itemize}
        \item ``№ Feats'' means Number of Features;
        \item ``Unselected'' means Features that should be selected but were not selected;
        \item ``Additional'' means Features that should not be selected are selected.
        \item Subsequent tables same as above.
    \end{itemize}
\end{table}

\subsubsection{Biased Sampling Results}
Considering that real world data may exhibit bias, the target function might be related to a certain feature, but samples possessing this feature may be scarce. We constructed a new dataset by randomly setting 99\% of the sample values in $X_{20}$ to $-1.0$ in the original dataset.

The test results indicate that neither the Boruta algorithm, nor Boruta-Permut, nor Boruta-TreeImp could select feature $X_{20}$, demonstrating that sample bias can have a significant effect on the Boruta feature selection procedure.

\subsubsection{Multicollinearity Results}
We constructed a new dataset based on the original one by adding five linearly correlated features that are highly related to the important features. The linearly correlated features were generated by adding Gaussian white noise to the original important features, $\epsilon_2 \sim \mathcal{N}(0, 0.01)$. The introduced variables are as follows:
\begin{equation}
    \begin{aligned}
        X_{13} &= 0.1X_{14} + 0.2X_{40} + 0.3X_{7} + 0.4X_{42} + \epsilon_2 \\
        X_{5} &= 0.4X_{31} + 0.2X_{46} + 0.3X_{47} + 0.1X_{48} + \epsilon_2 \\
        X_{38} &= 0.1X_{18} + 0.3X_{49} + 0.2X_{16} + 0.4X_{10} + \epsilon_2 \\
        X_{9} &= 0.4X_{27} + 0.3X_{26} + 0.2X_{17} + 0.1X_{25} + \epsilon_2 \\
        X_{4} &= 0.2X_{33} + 0.4X_{20} + 0.1X_{35} + 0.3X_{32} + \epsilon_2
    \end{aligned}
\end{equation}

Three algorithms were used to select features from the new dataset, and the results are shown in Table \ref{tab:simu_corr1}.
It can be seen that the Boruta algorithm, despite the introduction of linearly correlated features, still correctly selects all 10 important features but also includes 5 linearly correlated ones.
Compared to the standard Boruta algorithm, Boruta-Permut selects fewer linearly correlated features.
Boruta-TreeImp fails to select feature-18 and also includes features that are not linearly correlated with $X_{18}$, resulting in slightly poorer performance.

\begin{table}[!htb]
    \centering
    \caption{Comparison of Direct Selection Results on the Self-Constructed Dataset with multicollinearity Features}
    \label{tab:simu_corr1}
    \begin{tabular}{llllll}
        \toprule
        Algorithm & Time & Average Cost & № Feats & Unselected & Additional \\
        \midrule
        Boruta & 21m16s & 1.7300 USD & 10 & / & feature-4,5,9,38,47 \\
        Boruta-Permut & \textbf{03m36s} & \textbf{0.0071 USD} & 10 & / & feature-4,5,47 \\
        Boruta-TreeImp & 04m53s & 0.0096 USD & 9 & feature-18 & feature-4,5,47 \\
        \bottomrule
    \end{tabular}
\end{table}

\subsection{Experiments on Public Datasets}
We conducted experiments on the Relative location of CT slices on axial axis (CT) (\cite{Graf2011}), Online News Popularity (NEWS) (\cite{Fernandes2015}), Appliances Energy Prediction (ENERGY) (\cite{Candanedo2017}), and MADELON (\cite{Madelon2004}) datasets.
The MADELON is an artificially generated classification set with 500 samples, 20 useful features, and 480 noise features; the other datasets are regression sets whose features are all continuous.

We ran the original Boruta algorithm, the GPU-accelerated Boruta-Permut and Boruta-TreeImp algorithms, and the BorutaShap algorithm (\cite{BorutaShap}) for feature selection.
For each method, we performed 20 trials with different random seeds, recorded the feature importance scores in each trial, and used the median importance as the final selection.

Since MADELON provides a validation set, we built a random forest model on its training set and tested on the validation set using the selected features, recording classification accuracy.
The other datasets lack a validation set; for these, we used RandomForest as the base learner and applied 5-fold cross-validation on the selected features, reporting mean squared error (MSE), mean absolute error (MAE), and the coefficient of determination ($R^2$).

Because BorutaShap cannot run on classification tasks, it was not tested on MADELON.
The results are shown in Tables \ref{tab:select1} to \ref{tab:select4}.
It can be seen that, while ensuring selection effectiveness, the GPU-accelerated Boruta-Permut and Boruta-TreeImp algorithms significantly reduce costs and exhibit very high cost-effectiveness.
% 可以看出，在保证选择效果的前提下，GPU加速的Boruta-Permut和Boruta-TreeImp算法成本大幅降低，性价比极高。
\begin{table}[!htb]
    \centering
    \caption{Comparison of feature selection results on the CT dataset}
    \label{tab:select1}
    \begin{tabular}{lllllll}
        \toprule
        Algorithm & Time & Average Cost & № Feats & MSE & MAE & $R^2$ \\
        \midrule
        No Feature Select & / & / & 385 & \textbf{45.0526} & \textbf{3.0726}  & 0.9071  \\
        Boruta Origin & \textbf{26m00s} & 2.1151 USD & 146 & 45.0572  & 3.1283  & \textbf{0.9072}  \\
        Boruta-Shap & 05h20m03s & / & 172 & 47.4629  & 3.1811  & 0.9021 \\
        Boruta-Permut & 01h37m58s & 0.1927 USD & 123 & 48.0319  & 3.1744  & 0.9010  \\
        Boruta-TreeImp & 55m02s & \textbf{0.1082 USD} & 122 & 48.3319 & 3.2383 & 0.9005 \\
        \bottomrule
    \end{tabular}
\end{table}

\begin{table}[!htb]
    \centering
    \caption{Comparison of feature selection results on the NEWS dataset}
    \label{tab:select2}
    \begin{tabular}{lllllll}
        \toprule
        Algorithm & Time & Average Cost & № Feats & MSE & MAE & $R^2$ \\
        \midrule
        No Feature Select & / & / & 58 & $1.441\times10^8$  & 3553  & -0.0918  \\
        Boruta Origin & 21m44s & 1.7680 USD & 16 & $1.471\times10^8$  & 3512  & -0.1667  \\
        Boruta-Shap & 08h11m33s & / & 5 & $1.458\times10^8$  & \textbf{3504}  & -0.1241 \\
        Boruta-Permut & 13m59s & 0.0275 USD & 13 & $\mathbf{1.4259\times10^8}$  & 3509  & \textbf{-0.0735}  \\
        Boruta-TreeImp & \textbf{07m30s} & \textbf{0.0148 USD} & 5 & $1.460\times10^8$  & 3461  & -0.1126  \\
        \bottomrule
    \end{tabular}
\end{table}

\begin{table}[!htb]
    \centering
    \caption{Comparison of feature selection results on the ENERGY dataset}
    \label{tab:select3}
    \begin{tabular}{lllllll}
        \toprule
        Algorithm & Time & Average Cost & № Feats & MSE & MAE & $R^2$ \\
        \midrule
        No Feature Select & / & / & 27 & $2.482\times10^4$  & 116.8861  & -1.4181  \\
        Boruta Origin & 07m19s & 0.5952 USD & 23 & $2.342\times10^4$  & 113.8876  & -1.2810  \\
        Boruta-Shap & 07h38m24s & / & 23 & $2.342\times10^4$  & 113.8876  & -1.2810 \\
        Boruta-Permut & 15m54s & 0.0313 USD & 23 & $2.342\times10^4$  & 113.8876  & -1.2810  \\
        Boruta-TreeImp & \textbf{04m07s} & \textbf{0.0081 USD} & 7 & $\mathbf{1.791\times10^4}$  & \textbf{99.1073}  & \textbf{-0.7321}  \\
        \bottomrule
    \end{tabular}
\end{table}

\begin{table}[!htb]
    \centering
    \caption{Comparison of feature selection results on the MADELON dataset}
    \label{tab:select4}
    \begin{tabular}{llllllll}
        \toprule
        Algorithm & Time & Average Cost & № Feats & Accuracy & Recall & Precision & F1 Score \\
        \midrule
        No Feature Select & / & / & 500 & 0.6850  & 0.6800  & 0.6869  & 0.6834  \\
        Boruta Origin & 09m47s & 0.7959USD & 20 & 0.8883  & 0.9000  & 0.8795  & 0.8896  \\
        Boruta-Permut & 27m42s & 0.0545USD & 19 & \textbf{0.8983}  & \textbf{0.9067}  & \textbf{0.8918}  & \textbf{0.8992}  \\
        Boruta-TreeImp & \textbf{04m42s} & \textbf{0.0092USD} & 16 & 0.8917  & 0.9000  & 0.8852  & 0.8926 \\
        \bottomrule
    \end{tabular}
\end{table}

\begin{figure}[!htb]
    \centering
    \includegraphics[width=\textwidth]{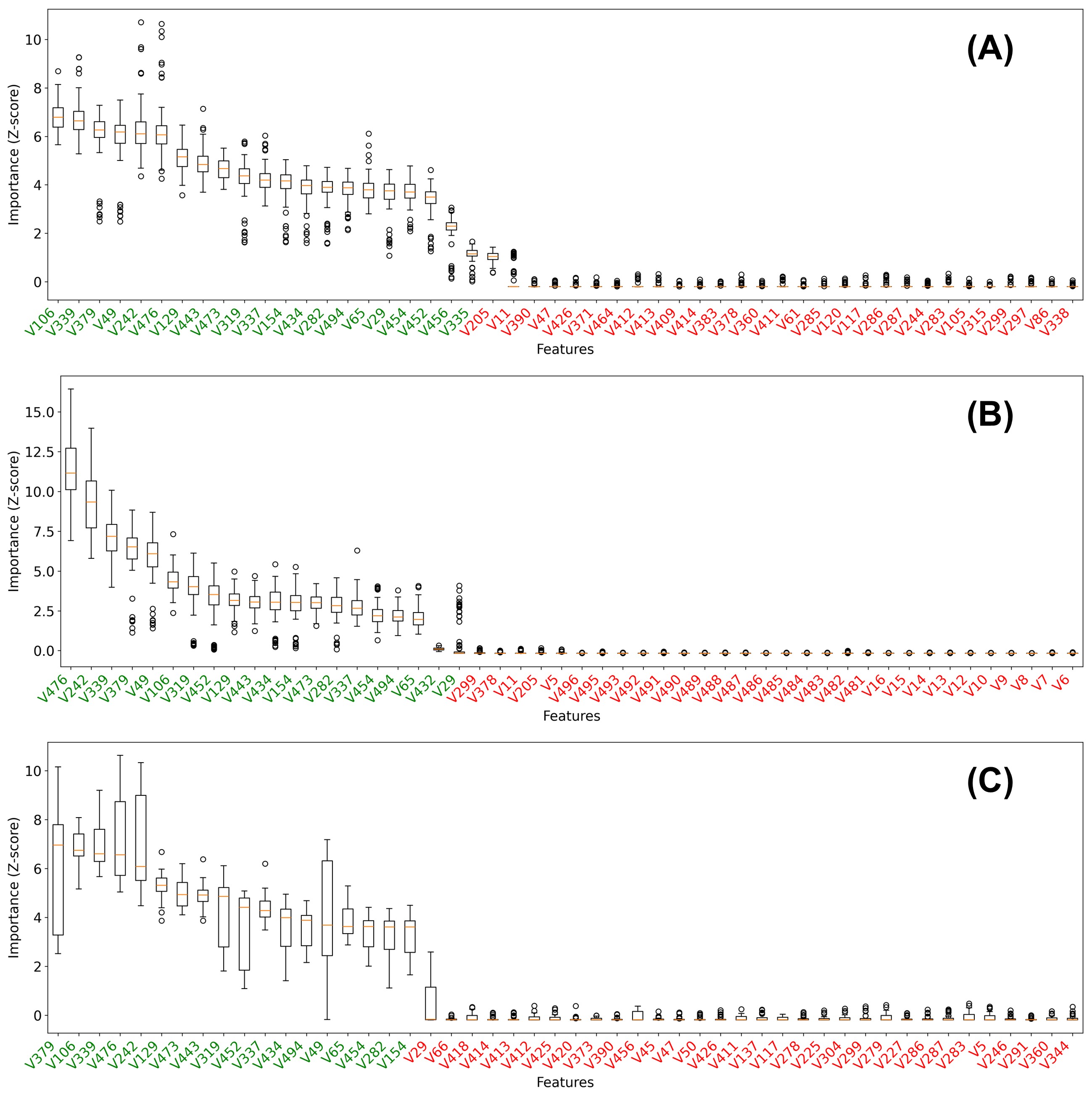}
    \caption{
        \textbf{Comparison of feature selection results on the MADELON dataset}
        \\
        (A) Boruta; (B) Boruta-Permut; (C) Boruta-TreeImp.
        \\
        Green indicates that a feature is accepted; blue indicates that it is tentative; red indicates that it is rejected
    }
    \label{fig:simu_direct2}
\end{figure}

\begin{figure}[!htb]
    \centering
    \includegraphics[width=\textwidth]{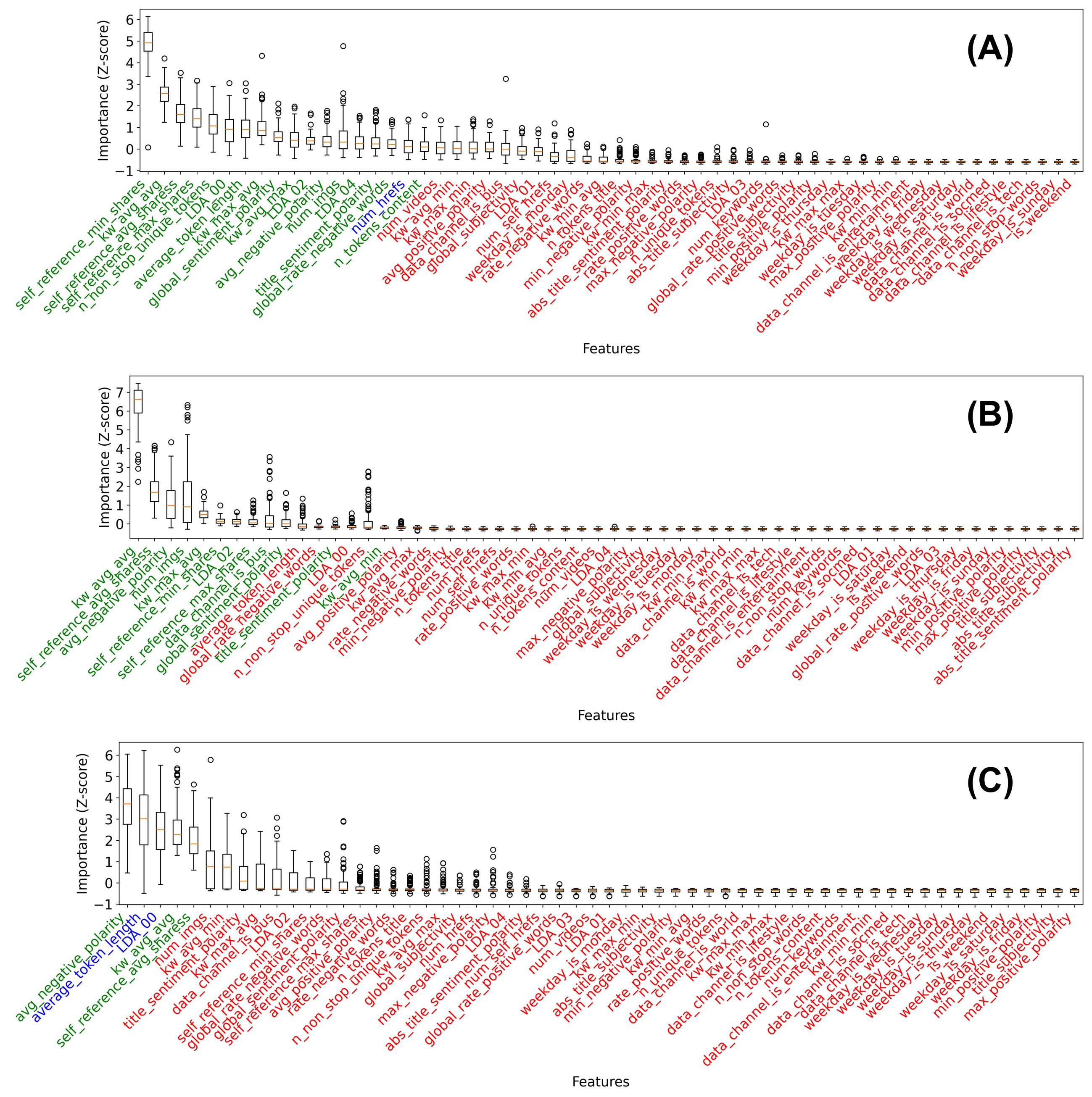}
    \caption{
        \textbf{Comparison of feature selection results on the NEWS dataset}
        \\
        (A) Boruta; (B) Boruta-Permut; (C) Boruta-TreeImp.
        \\
        Green indicates that a feature is accepted; blue indicates that it is tentative; red indicates that it is rejected
    }
    \label{fig:simu_direct3}
\end{figure}

Figures \ref{fig:simu_direct2} and \ref{fig:simu_direct3} show the comparison of the selection results of the three algorithms on MADELON and NEWS, respectively.
For the MADELON, all algorithms provided relatively consistent selection results, although there were differences in the distribution of feature importance.
However, for the NEWS dataset, Boruta-TreeImp selected significantly fewer features than the other two algorithms, which may be related to the tendency of impurity-based feature importance to overestimate the importance of shadow features.
% 对于Madelon数据集，各个算法给出了较为解决的选择结果，不过特征重要性的分布存在差异。
% 但对于Online News Popularity数据集，Boruta-TreeImp选出的特征数量远少于其他两种算法，这可能与impurity-based feature importance夸大了Shadow features的重要性有关。

\section{Conclusion and Future Outlook}
Using cuML, we accelerated the Boruta algorithm on GPUs and proposed two variants: Boruta-Permut and Boruta-TreeImp.
We tested these algorithms on both a self-constructed dataset and public datasets, and compared them with the Boruta and Boruta-Shap implementations in scikit-learn.
The experimental results show that our methods significantly improve computational efficiency while maintaining feature-selection effectiveness, indicating promising application potential.

This work propose two GPU accelerated Boruta algorithm and demonstrated their effectiveness in feature selection and computational economics in processing large datasets. The experiments on the self-constructed dataset indicate that Boruta-TreeImp may miss important features when dealing with certain complex functions, which is related to the limitations of impurity-based feature importance potentially overestimating the importance of irrelevant features.
Enhancing the algorithm's scalability will be included into the future study.
Firstly, implementing multi-GPU parallelism within a distributed computing framework for ultra-large datasets will be tested.
Secondly, new feature importance calculation methods should be attempted to improve the computational efficiency of Boruta-Permut while reduce the missing of important features when using Boruta-TreeImp.
% Using cuML, we accelerated the Boruta algorithm on GPUs and proposed two variants: Boruta-Permut and Boruta-TreeImp.
% We tested these algorithms on both a self-constructed dataset and public datasets, and compared them with the Boruta and Boruta-Shap implementations in scikit-learn.
% The experimental results show that our methods significantly improve computational efficiency while maintaining feature-selection effectiveness, indicating promising application potential.

% This work expands the application scope of the Boruta algorithm and demonstrates its cost advantage when processing large datasets.
% Compared to the original algorithm, the GPU-accelerated version offers substantial benefits.

% However, experiments on the self-constructed dataset indicate that Boruta-TreeImp may miss important features when dealing with certain complex functions, which is related to the limitations of impurity-based feature importance potentially overestimating the importance of irrelevant features.

% Future work may focus on further enhancing the algorithm's scalability.
% Firstly, implementing multi-GPU parallelism within a distributed computing framework could further improve its performance on ultra-large datasets.
% Secondly, developing new methods for calculating feature importance could be a focus, as Boruta-Permut currently has slightly lower computational efficiency, and Boruta-TreeImp may miss important features; future efforts could aim to create new feature importance calculation methods to enhance the overall performance of the algorithm.

\section*{Data availability}
All of the public datasets were downloaded from the UCI Machine Learning Repository.
And the references are listed in the article.

\section*{Code availability}
The complete code has been released on Github:
\href{https://github.com/A-Normal-User/GPUBoruta}{https://github.com/A-Normal-User/GPUBoruta}

\section*{Acknowledgements}
This work was supported by the National Natural Science Foundation of China, Grand No.U24A6011 and by Tsinghua University No. 2023Z020RD001.

\section*{Author information}
\subsection*{Authors and Affiliations}
\noindent
\textbf{Department of Chemical Engineering, Tsinghua University, Beijing, China.}\\
Xurui Li, Zhiguo Gan, Jiaming Zhang, Diannan Lu, Zheng Liu.

\subsection*{Contributions}
\noindent
Zhiguo Gan and Jiaming Zhang conducted the preliminary research; Xurui Li implemented the code, carried out the experiments, and drafted the manuscript; Zhiguo Gan, Diannan Lu, and Zheng Liu reviewed and validated the manuscript.

\subsection*{Corresponding author}
\noindent
Correspondence to Diannan Lu and Zheng Liu.
% Some journals require declarations to be submitted in a standardised format. Please check the Instructions for Authors of the journal to which you are submitting to see if you need to complete this section. If yes, your manuscript must contain the following sections under the heading `Declarations':

% \begin{itemize}
% \item Conflict of interest/Competing interests (check journal-specific guidelines for which heading to use)
% \item Ethics approval and consent to participate
% \item Consent for publication
% \item Materials availability
% \end{itemize}

% \noindent
% If any of the sections are not relevant to your manuscript, please include the heading and write `Not applicable' for that section. 

% %%===================================================%%
% %% For presentation purpose, we have included        %%
% %% \bigskip command. Please ignore this.             %%
% %%===================================================%%
% \bigskip

% Springer journals and proceedings: \url{https://www.springer.com/gp/editorial-policies}

% Nature Portfolio journals: \url{https://www.nature.com/nature-research/editorial-policies}

% \textit{Scientific Reports}: \url{https://www.nature.com/srep/journal-policies/editorial-policies}

% BMC journals: \url{https://www.biomedcentral.com/getpublished/editorial-policies}
% \end{flushleft}

\begin{appendices}
% \section{Code Available}\label{secA1}
% The subsequent code will be published on GitHub in the future.

% An appendix contains supplementary information that is not an essential part of the text itself but which may be helpful in providing a more comprehensive understanding of the research problem or it is information that is too cumbersome to be included in the body of the paper.




\end{appendices}


\bibliography{ref}% common bib file

\end{document}